\documentclass{article}

\usepackage{microtype}
\usepackage{graphicx}
\usepackage{subfig}
\usepackage{booktabs} 
\usepackage{multirow}
\usepackage{makecell}
\usepackage{amsfonts}
\usepackage{amsmath}
\usepackage{bbold}

\usepackage{hyperref}

\usepackage[accepted]{icml2021}

\icmltitlerunning{Probabilistic Time Series Forecasting with Implicit Quantile Networks}

\begin{document}

\twocolumn[
\icmltitle{Probabilistic Time Series Forecasting with Implicit Quantile Networks}

\icmlsetsymbol{equal}{*}

\begin{icmlauthorlist}
\icmlauthor{Ad\`ele Gouttes}{equal,zalando}
\icmlauthor{Kashif Rasul}{equal,zalando}
\icmlauthor{Mateusz Koren}{zalando}
\icmlauthor{Johannes Stephan}{zalando}
\icmlauthor{Tofigh Naghibi}{zalando}
\end{icmlauthorlist}
\icmlaffiliation{zalando}{Zalando SE, Valeska-Gert-Straße 5, 10243 Berlin, Germany}
\icmlcorrespondingauthor{Adele Gouttes}{adele.gouttes@zalando.de}
\icmlkeywords{Time Series Workshop, ICML}

\vskip 0.3in
]

\printAffiliationsAndNotice{\icmlEqualContribution}

\begin{abstract}
    Here, we propose a general method for probabilistic time series forecasting. We combine an autoregressive recurrent neural network to model temporal dynamics with Implicit Quantile Networks to learn a large class of distributions over a time-series target. When compared to other probabilistic neural forecasting models on real- and simulated data, our approach is favorable in terms of point-wise prediction accuracy as well as on estimating the underlying temporal distribution.
\end{abstract}


\section{Introduction}

Despite being versatile and omnipresent, traditional time series forecasting methods such as those in~\cite{hyndman2018forecasting}, typically provide univariate point forecasts. Training in such frameworks requires to learn one model per individual time series, which might not scale for large data. Global deep learning-based time series models are typically recurrent neural networks (RNN), like  LSTM~\citep{6795963}. These methods have become popular due to their end-to-end training, the ease of incorporating exogenous covariates, and their automatic feature extraction abilities, which are the hallmarks of deep learning. 


It is often desirable for the outputs to be probability distributions, in which case forecasts typically provide uncertainty bounds.
In the deep learning setting the two main approaches to estimate uncertainty have been to either model the data distribution explicitly or to use Bayesian Neural Networks as in \cite{8215650}. The former methods rely on some parametric density function, such as that of a Gaussian distribution, which is often  based on computational convenience rather than on the underlying distribution of the data.

In this paper, we propose \texttt{IQN-RNN}, a deep-learning-based univariate time series method that learns an implicit distribution over outputs. Importantly, our approach does not make any a-priori assumptions on the underlying distribution of our data. The probabilistic output of our model is generated via Implicit Quantile Networks \cite{pmlr-v80-dabney18a} (IQN) and is trained by  minimizing the integrand of the Continuous Ranked Probability Score (CRPS) \cite{RePEc}.

The major contributions of this paper are:
\begin{enumerate}
    \item  model the data distribution using IQNs which allows the use of a broad class of datasets;
    \item  model the time series via an autoregressive RNN  where the emission distribution is given by an IQN;
    \item demonstrate competitive results on real-world datasets in particular when compared to RNN-based probabilistic univariate time series models.
\end{enumerate}

\section{Background}\label{sec:background}

\subsection{Quantile Regression}

The quantile function corresponding to a cumulative distribution function (c.d.f.) $F \colon \mathbb{R} \to [0,1]$ is defined as:
\begin{equation*}
    Q(\tau) = \inf\{x\in\mathbb{R}: \tau \leq F(x)\}.
\end{equation*}
For continuous and strictly monotonic  c.d.f. one can simply write $Q = F^{-1}$.

In order to find the quantile for a given $\tau$ one can use quantile regression \cite{koenker_2005}, which minimizes the quantile loss:
\begin{equation}
    L_\tau(y,\hat{y}) = \tau(y - \hat{y})_+ + (1-\tau)(\hat{y} - y)_+,
\label{eq:quantile-loss}
\end{equation}
where $()_+$ is non-zero iff the value in the parentheses is positive.


\subsection{CRPS}
Continuous Ranked Probability Score \cite{RePEc,Gneiting_2007} is a \emph{proper} scoring rule, described by a c.d.f. $F$ given the observation $y$:
\begin{equation*}
    \textrm{CRPS}(F, y) = \int_{-\infty}^{\infty}(F(x) - \mathbb{1}\{y\leq x\})^2 \,dx,
\end{equation*}
where $\mathbb{1}$ is the indicator function. The formula can be rewritten \cite{hess-11-1267-2007} as an integral over the quantile loss:
\begin{equation*}
    \textrm{CRPS}(F, y) = 2\int_{0}^{1} L_{\tau}(y, Q(\tau)) \,d\tau,
\end{equation*}
where $Q$ is the quantile function corresponding to $F$.

\section{Related work}


Over the last years, deep learning models have shown impressive results over classical methods in many fields~\cite{888} like computer vision, speech recognition, natural language processing (NLP), and also time series forecasting, which is related to sequence modeling in NLP~\cite{NIPS2014_5346}. Modern univariate forecast methods like \texttt{N-BEATS}~\cite{oreshkin2020nbeats} share parameters, are interpretive and fast to train on many target domains.  

 To estimate the underlying temporal distribution we can learn the parameters of some target distribution as in the \texttt{DeepAR} method \cite{DBLP:journals/corr/FlunkertSG17} or use mixture density models~\cite{McLa1988} operating on neural network outputs, called mixture density networks (MDN)~\cite{bishop:2006:PRML}. One prominent example is \texttt{MDRNN}~\cite{Graves13} that uses a mixture-density RNN to model handwriting. These approaches assume some parametric distribution, based on the data being modeled, for example a Negative Binomial distribution for count data. Models are trained by maximizing the likelihood of these distributions with respect to their predicted parameters and ground truth data. 
 
 Our approach is closely related to the \texttt{SQF-RNN} \cite{pmlr-v89-gasthaus19a} which models the conditional quantile function using isotonic splines. We utilize an IQN \cite{pmlr-v80-dabney18a, yang2019fully} instead, as we will detail, which has been used in the context of Distributional Reinforcement Learning \cite{pmlr-v70-bellemare17a}, as well as for Generative Modelling \cite{pmlr-v80-ostrovski18a}.

\section{Forecasting with Implicit Quantile Networks}

In an univariate time series forecasting setting, we typically aim at forecasting a subseries $(y_{T-h}, y_{T-h+1}, \dots, y_T)$ of length $h$ from a series $y = (y_0, y_1, \dots, y_T)$ of length $T + 1$, generated by an auto-regressive process $Y=(Y_0, Y_1, \dots, Y_T)$. 

Let $\tau_0 = \mathbb{P}[Y_0 \leq y_0]$ and $\tau_t = \mathbb{P}[Y_t \leq y_t | Y_0 \leq y_0, \ldots, Y_{t-1} \leq y_{t-1}]$ for each integer $t \in [\![1, T ]\!]$. We can rewrite $y$ as $(F_{Y_0}^{-1}(\tau_0), F_{Y_1 | Y_0}^{-1}(\tau_1), \dots, F_{Y_T | Y_0, \dots, Y_T}^{-1}(\tau_T))$, where $F$ is the c.d.f.. 

In probabilistic time series forecasting, it is typically assumed that a unique function $g$ can represent the distribution of all $Y_t$, given  input covariates $X_t$ and previous observations $y_0, y_1, \dots, y_{t-1}$. When using IQNs, we additionally parameterize this function with $\tau_t$: IQNs learn the mapping from $\tau_t \sim \ \textrm{U}([0,1])$ to $y_t$. In other words, they are deterministic parametric functions trained to reparameterize samples from the uniform distribution to respective quantile values of a target distribution. Our \texttt{IQN-RNN} model should then learn $y_t = g(X_t, \tau_t, y_0, y_1, \dots, y_{t-1})$ for $t \in [\![T-h, T ]\!]$ and can be written as $q \circ [\psi_t \odot (1 + \phi)]$, where $\odot$ is the Hadamard element-wise product and:
\begin{itemize}
    \item $X_t$ are typically time-dependent features, known for all time steps;
    \item $\psi_t$ is the state of an RNN that takes the concatenation of $X_t$, $y_{t-1}$ as well as the previous state $\psi_{t-1}$ as input;
    \item $\phi$ embeds a $\tau_t$ as described by \cite{pmlr-v80-dabney18a}, with $n=64$:
\begin{equation*}
    \phi(\tau_t) = \textrm{ReLU}(\sum_{i=0}^{n-1}\cos(\pi i \tau_t)w_i + b_i);
\end{equation*}
    \item $q$ is an additional generator layer, which in our case is a simple two-layer feed-forward neural network, with a domain relevant activation function.
\end{itemize}
We perform the Hadamard operation on $(1+ \phi)$, which is one of the forms considered by \cite{pmlr-v80-dabney18a}.

\begin{figure}[ht]
\vskip 0.2in
\begin{center}
\centerline{\includegraphics[width=\columnwidth]{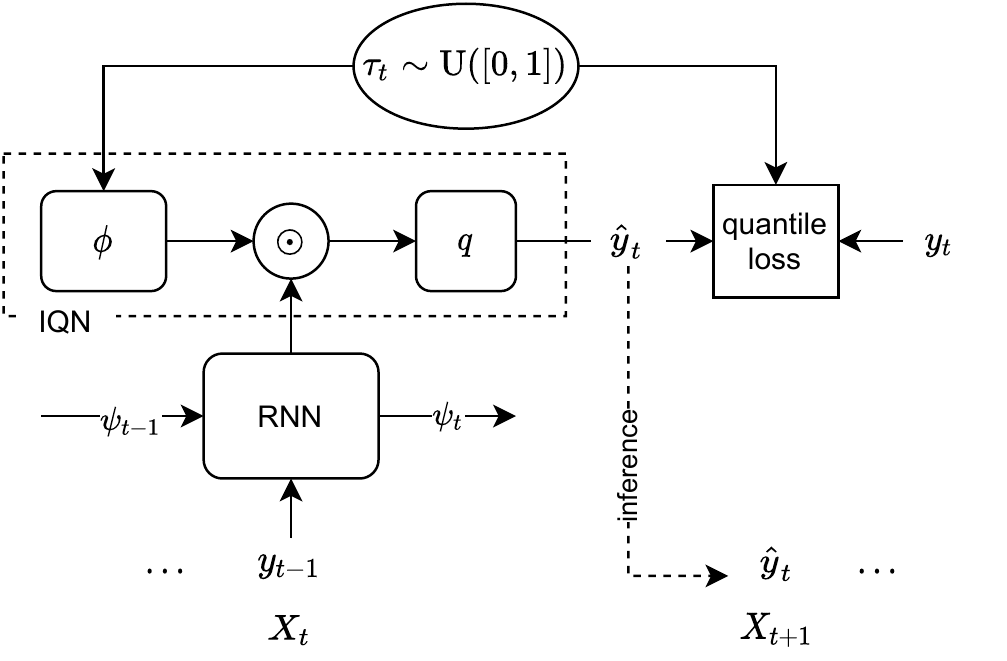}}
\caption{\texttt{IQN-RNN} schematic at time $t$ where during training we minimize the quantile loss with respect to the ground truth.}
\label{time-grad-fig}
\end{center}
\vskip -0.2in
\end{figure}

At training time, quantiles are sampled for each observation at each time step and passed to both network and quantile loss (\ref{eq:quantile-loss}) (see Figure~\ref{time-grad-fig}). 

During inference, we analogously sample a new quantile for each time step of our autoregressive loop. Thus, a full single forward pass follows an \emph{ancestral sampling} scheme along the graph of our probabilistic network. This approach guarantees to produce valid samples from the underlying model. Sampling a larger number of  trajectories this way, allows us to estimate statistics over the distribution of each observation such as mean, quantiles, and standard deviation. For instance, the mean of $Y_{T-h}$ can be estimated using the average over the first step of all sampled trajectories. In our experiments, we choose 100 samples (in parallel via the batch dimension) when calculating metrics and empirical quantiles. This  strategy also addresses potential quantile-crossing issues, since nothing in the \texttt{IQN-RNN} architecture guaranties monotonicity with respect to $\tau$: we simply compute the quantiles from sampled values.

We note that this method would work equally well using a masked Transformer \cite{NIPS2017_7181} to model the temporal dynamics or a fixed horizon non-autoregessive model like in  \cite{oreshkin2020nbeats}.

\section{Experiments}

We evaluate our \texttt{IQN-RNN} model on synthetic and open  datasets and follow the recommendations of the M4 competition \cite{MAKRIDAKIS202054} regarding performance metrics. We report the mean scale interval score (MSIS\footnote{\url{https://bit.ly/3c7ffmS}}) for a 95\% prediction interval, the 50-th and 90-th quantile percentile loss, and the CRPS. The point-wise performance of models is measured by the normalized root mean square error (NRMSE), the mean absolute scaled error (MASE) \cite{HYNDMAN2006679}, and the symmetric mean absolute percentage error (sMAPE) \cite{MAKRIDAKIS1993527}. For pointwise metrics, we use sampled medians with the exception of NRMSE, where we take the mean over our prediction sample.

The code for our model is available in the \texttt{PyTorchTS}~\cite{pytorchgithub} library.

\subsection{Results on synthetic data}

We firstly evaluate our \texttt{IQN-RNN} on synthetic data and compare its performance with another non-parametric probabilistic forecast model: \texttt{SQF-RNN} \cite{pmlr-v89-gasthaus19a}. In order to minimize the MSIS and CRPS, we use this model with 50 linear pieces. Both models have the same \texttt{RNN} architecture and the same hyperparameters for training. Only the probabilistic head is distinct.

In a similar fashion to \cite{pmlr-v89-gasthaus19a}, we generate 10,000 time series of 48 points, where each time step is $iid$ and follows a Gaussian Mixture Model: $GMM(\pi, \mu, \sigma)$ with $\pi=[.3, .4, .3]^T$, $\mu=[-3, 0, 3]^T$, $\sigma=[.4, .4, .4]^T$. The models are trained 5 times each, with a context length of 15, a prediction window of 2 for 20 epochs. 

\begin{figure}[t]
    \centering
    \subfloat[\centering \texttt{IQN-RNN}]{\includegraphics[width=.87\columnwidth]{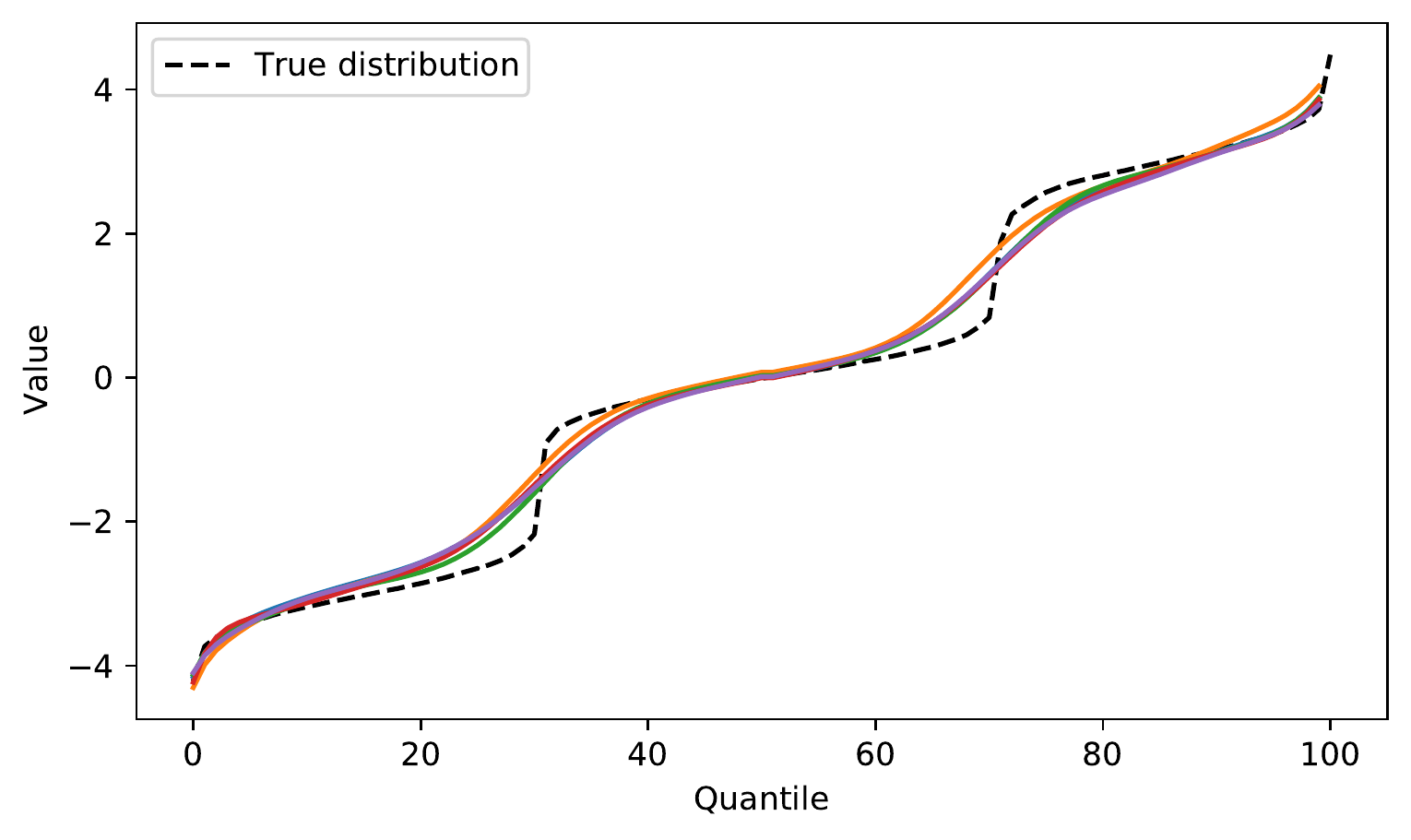} }%
    \qquad
    \subfloat[\centering \texttt{SQF-RNN}]{{\includegraphics[width=.87\columnwidth]{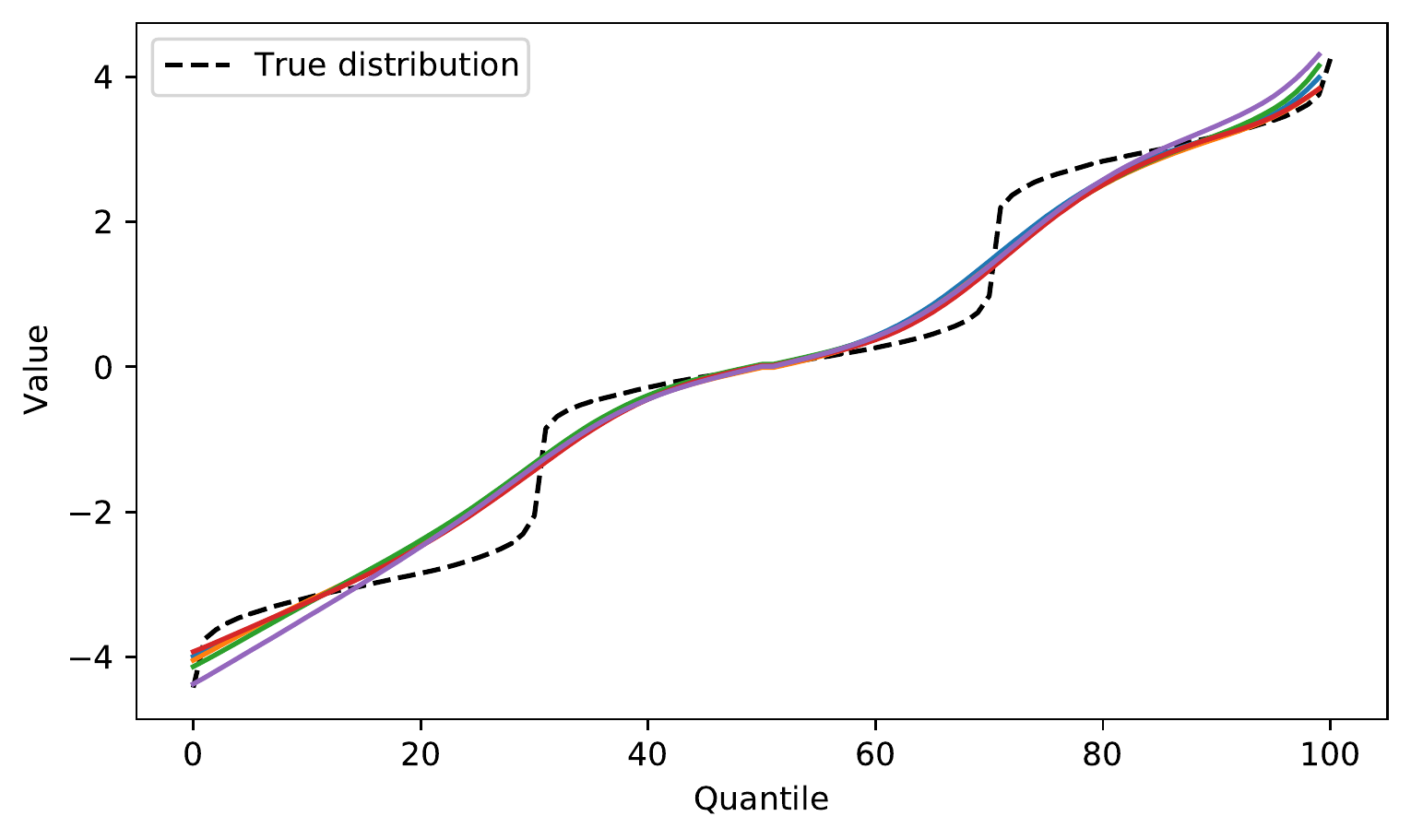} }}%
    \caption{Estimated quantile functions for five training on time series following a Gaussian Mixture  using (a) \texttt{IQN-RNN} and (b) \texttt{SQF-RNN} model.}%
    \label{estimated-quantiles}%
\end{figure}

We show the estimated quantile functions in Figure~\ref{estimated-quantiles} and report the average metrics in Table~\ref{synthetic-metrics}. While both methods have a similar point-wise performance, \texttt{IQN-RNN} is better at estimating the entire probability distribution. 

\begin{table}[t]
\caption{Performance of \texttt{IQN-RNN} and \texttt{SQF-RNN} in fitting a three-component Gaussian Mixture Model.}
\label{synthetic-metrics}
\vskip 0.15in
\begin{center}
\begin{small}
\begin{sc}
\begin{tabular}{lccccccr}
\toprule
Method & CRPS & MSIS & sMAPE & MASE \\
\midrule
\texttt{SQF-RNN-50}  & 0.780 & 3.213 & 1.756 & 0.747  \\
\texttt{\textbf{IQN-RNN}}  & \textbf{0.776}  & \textbf{3.027} &  1.754 & 0.740 \\

\bottomrule
\end{tabular}
\end{sc}
\end{small}
\end{center}
\vskip -0.1in
\end{table}

\subsection{Results on empirical data}

\begin{table*}[t]
\caption{Comparison against different methods: \texttt{SQF-RNN} with 50 nodes, \texttt{DeepAR} with Student-T (\texttt{-t}) or Negative Binomial (\texttt{-nb}) output, \texttt{ETS} and \texttt{IQN-RNN} on the 3 datasets.}
\label{metrics}
\vskip 0.15in
\begin{center}
\begin{small}
\begin{sc}
\begin{tabular}{lcccccccr}
\toprule
Data set & Method & CRPS & QL50 & QL90 & MSIS & NRMSE & sMAPE & MASE \\
\midrule
\multirow{4}{*}{\texttt{Electricity}} 
  & \texttt{SQF-RNN-50}  & 0.078 & 0.097 & 0.044 &   8.66  &0.632 & 0.144 & 1.051 \\ 
  & \texttt{DeepAR-t}  & 0.062 & 0.078 & 0.046 &  \textbf{6.79} & 0.687 & 0.117 & \textbf{0.849} \\ 
  & \texttt{ETS}  & 0.076 & 0.100 & 0.050 & 9.992 & 0.838 & 0.156 & 1.247  \\
  & \texttt{\textbf{IQN-RNN}}  & \textbf{0.060}  & \textbf{0.074} & \textbf{0.040} &  8.74 &  \textbf{0.543} & \textbf{0.138} & 0.897 \\ 
\hline

\multirow{4}{*}{\texttt{Traffic}} 
 & \texttt{SQF-RNN-50} & 0.153 & 0.186 & \textbf{0.117} & 8.40 & \textbf{0.401} &  0.243  & 0.76 \\ 
 & \texttt{DeepAR-t} & 0.172 & 0.216 & \textbf{0.117} & 8.027 & 0.472  & 0.244 &0.89\\ 
 & \texttt{ETS} & 0.427 & 0.488 & 0.325 & 20.856 & 0.872 & 0.594 & 1.881 \\
 & \texttt{\textbf{IQN-RNN}}  & \textbf{0.139} & \textbf{0.168} & \textbf{0.117} & \textbf{7.11} & 0.433 & \textbf{0.171} & \textbf{0.656} \\
\hline

\multirow{4}{*}{\texttt{Wikipedia}}
   & \texttt{SQF-RNN-50} &0.283 & 0.328 & 0.321 & 23.71 &  2.24  & 0.261 & 1.44 \\ 
   & \texttt{DeepAR-nb}  &0.452  & 0.572 &  0.526 & 46.79 & 2.25 & 0.751  & 2.94 \\ 
   & \texttt{DeepAR-t}  & 0.235 &0.27 &  0.267 & 23.77 & 2.15 & 0.21  & 1.23 \\ 
   & \texttt{ETS} & 0.788 & 0.440 & 0.836 & 61.685 & 3.261 & 0.301 & 2.214 \\
   & \texttt{\textbf{IQN-RNN}}  & \textbf{0.207} & \textbf{0.241} & \textbf{0.238}
   &  \textbf{19.61} & \textbf{2.074} & \textbf{0.179} &  \textbf{1.141} \\
\bottomrule
\end{tabular}
\end{sc}
\end{small}
\end{center}
\vskip -0.1in
\end{table*}

\begin{table}[hbt]
\caption{Number of time series, domain, frequency, total training time steps and prediction length properties of the  training datasets used in the experiments.}
\label{dataset}
\vskip 0.15in
\begin{center}
\begin{small}
\begin{sc}
\begin{tabular}{lccccr}
\toprule
Data set &  Num. & Dom. & Freq. & \makecell{Time \\ steps}  & \makecell{Pred.\\ steps}  \\
\midrule
 \texttt{Elec.} & $321$ &  $\mathbb{R}^{+}$ & hour & $21,044$ & $24$  \\
\texttt{Traffic}  & $862$  & $(0,1)$ & hour & $14,036$ & $24$  \\
\texttt{Wiki.}   & $9,535$ & $\mathbb{N}$ & day & $762$ & $30$ \\
\bottomrule
\end{tabular}
\end{sc}
\end{small}
\end{center}
\vskip -0.1in
\end{table}

\begin{table}[hbt]
\caption{Common hyperparmeters for \texttt{SQF-RNN}, \texttt{DeepAR} and \texttt{IQN-RNN} models.}
\label{hyper-params}
\vskip 0.15in
\begin{center}
\begin{small}
\begin{sc}
\begin{tabular}{lr}
\toprule
Hyperparameter &  Value  \\
\midrule
\texttt{rnn\_cell\_type} & GRU\cite{69e088c8129341ac89810907fe6b1bfe}  \\
\texttt{rnn\_hidden\_size}  & 64  \\
\texttt{rnn\_num\_layers}   & 3 \\
\texttt{rnn\_dropout\_rate}   & 0.2 \\
\texttt{context\_length}   & \texttt{2 * pred\_steps} \\
\texttt{epochs}   & 10 \\
\texttt{learning\_rate}   & 0.001 \\
\texttt{batch\_size}   & 256 \\
\texttt{batches\_per\_epoch}   & 120 \\
\texttt{num\_samples}   & 100 \\
\texttt{optim} & Adam \cite{kingma:adam} \\
\bottomrule
\end{tabular}
\end{sc}
\end{small}
\end{center}
\vskip -0.1in
\end{table}

We next evaluate our model on open source datasets for univariate time series: \texttt{Electricity}\footnote{\url{https://archive.ics.uci.edu/ml/datasets/ElectricityLoadDiagrams20112014}}, \texttt{Traffic}\footnote{\url{https://archive.ics.uci.edu/ml/datasets/PEMS-SF}}, and \texttt{Wikipedia}\footnote{\url{https://github.com/mbohlkeschneider/gluon-ts/tree/mv_release/datasets}}, preprocessed exactly as in~\cite{NIPS2019_8907}, with their properties listed in Table~\ref{dataset}.
Our model is trained on the training split of each dataset. For testing, we use a rolling-window prediction starting from the last point seen in the training dataset and compare it to the test set.

For comparison, we again use \texttt{SQF-RNN} \cite{pmlr-v89-gasthaus19a} with 50 linear pieces. We also evaluate \texttt{DeepAR} \cite{DBLP:journals/corr/FlunkertSG17} with a Student-T or a Negative Binomial distribution depending on the domain of the dataset. Since \texttt{IQN-RNN}, \texttt{SQF-RNN} and \texttt{DeepAR} share the same RNN architecture  we compare these models using the same \emph{untuned}, but recommended, hyperparameters (see Table \ref{hyper-params}) for training: only the probabilistic heads differ. Thus, deviations from performance reported in the original publications are solely due to the number of epochs used for training. Alternative models are trained on the same instances, consume a similar amount of memory, and need  similar training time. We also use \texttt{ETS} \cite{JSSv027i03} as a comparison, which is an exponential smoothing method using weighted averages of past observations with exponentially decaying weights as the observations get older together with Gaussian additive errors (E) modeling trend (T) and seasonality (S) effects separately. 

In Table~\ref{metrics} we report  probabilistic and point-wise performance metrics of all models. We found that using \texttt{IQN-RNN} often leads to the best performance on both probabilistic and point-wise metrics while being fully non-parametric and without significantly increasing the parameters of the \texttt{RNN} model. We also note, that the resulting performance on point-forecasting metrics does not result in higher errors for our probabilistic measures (unlike e.g. \texttt{DeepAR}). We did not incorporate per time series embeddings as covariates in any of our experiments. 

\section{Conclusion}

In this work, we proposed a general method of probabilistic time series forecasting by using IQNs to learn the quantile function of the next time point. We demonstrated the performance of our approach against competitive probabilistic methods on real-world datasets.

Our framework can be easily extended to multivariate time series, under the rather restrictive hypothesis that we observe the same quantile for individual univariate series. This is equivalent to assuming \emph{comonotonicity} of the processes for each time step. Relaxing this assumption is left to future research.

\section*{Software}

We wish to acknowledge and thank the authors and contributors of the following open source libraries that were used in this work: GluonTS ~\citep{gluonts_jmlr}, NumPy~\citep{harris2020array}, Pandas~\citep{reback2020pandas}, Matplotlib~\citep{Hunter:2007} and  PyTorch~\citep{NIPS2019_9015}. 

\section*{Acknowledgements}

K.R.: I acknowledge the traditional owners of the land on which I have lived and worked, the Wurundjeri people of the Kulin nation who have been custodians of their land for thousands of years. I pay my respects to their elders, past and present as well as past and present aboriginal elders of other communities.

\bibliography{references}
\bibliographystyle{icml2021}
\end{document}